\documentclass[letterpaper, 10 pt, conference]{ieeeconf}
\IEEEoverridecommandlockouts                              % This command is only
\usepackage[utf8]{inputenc}
\usepackage[T1]{fontenc}
\usepackage{csquotes}
\usepackage{color}
\usepackage[pdftex,dvipsnames]{xcolor}
\usepackage{graphicx}
\usepackage{hyperref}
\usepackage{xargs}
\usepackage{wrapfig}
\usepackage{textcomp}
\usepackage{bm}
\usepackage{amsmath}
\usepackage{mathtools}
\usepackage{gensymb}
\usepackage{booktabs}
\usepackage{adjustbox}
\usepackage{float}
\usepackage{helvet}
\usepackage{amsfonts}

\usepackage{tikz}
\usepackage{xfrac}
\usepackage{tabu} %https://tex.stackexchange.com/questions/10535/how-to-force-a-table-into-page-width

\usepackage[ruled,vlined]{algorithm2e}

\newcommand{\etal}{\textit{et al.}}
\newcommand{\cmt}[1]{\ignorespaces}

\newcommand{\HBFN}[1]{HB#1-FN}

\newcommand{\nDataset}{five}
\newcommand{\cTimeRatio}{50\%}

\newcommand{\bestD}{12}
\newcommand{\HBbest}{\HBFN{\bestD{}}}
\newcommand{\bestT}{9.1\,ms}
\newcommand{\bestEP}{0.553}
\newcommand{\HOGEP}{0.318}
\newcommand{\HOGtpGAP}{28.6\%}
\newcommand{\HOGtimeGAP}{7\,ms}

\newcommand{\vMargin}{\vspace{2mm}}

\newcommand{\fullpath}[1]{./img/#1}

\usepackage{tikz}

\newcommand*\annotatedFigureText[4]{\node[draw=none, anchor=south west, text=#2, inner sep=0, text width=#3\linewidth,font=\sffamily] at (#1){#4};}

\newenvironment {annotatedFigure}[1]{\centering\begin{tikzpicture}
    \node[anchor=south west,inner sep=0] (image) at (0,0) { #1};\begin{scope}[x={(image.south east)},y={(image.north west)}]}{\end{scope}\end{tikzpicture}}
\newcommand{\perfS}{EP}
\newcommand{\rp}{R_{P100}}
\newcommand{\pr}{P_{R0}}

\newcommand{\tf}{T_{i}}
\newcommand{\powerW}{P_w}

\newcommand{\energy}{E_i}

\newcommand{\art}{state-of-the-art}

\newcommand{\HB}{HB-DS}

\usepackage{multirow}
\usepackage{bigstrut}
\usepackage{makecell}

\newcommand{\bferra}{Bruno Ferrarini}

\newcommand{\bferraEm}{bferra}

\newcommand{\sheh}{Shoaib Ehsan}

\newcommand{\shehEm}{sehsan}

\newcommand{\kdm}{Klaus D. McDonald-Maier}

\newcommand{\kdmEm}{kdm}

\newcommand{\mm}{Michael Milford}
\newcommand{\mmEM}{michael.milford@qut.edu.au}

\title{\LARGE \bf
Highly-Efficient Binary Neural Networks for Visual Place Recognition
}

\author{\bferra{}$^{1}$, \mm{}$^{2}$, \kdm{}$^{1}$ and \sheh{}$^{1}$ % <-this % stops a space
\thanks{This work was supported by the UK Engineering and Physical Sciences
Research Council through grants EP/R02572X/1, EP/P017487/1, and in part by the RICE project funded by the National Centre for Nuclear Robotics Flexible Partnership Fund.}% <-this % stops a space
\thanks{$^{1}$\bferra{}, \kdm{} and \sheh{} are with the School of Computer Science and Electronic Engineering, University of Essex, Colchester, CO4 3SQ, UK
        {\tt\small \{\bferraEm{},\kdmEm{}, \shehEm{}\}@essex.ac.uk}}%
\thanks{\mm{} is with the QUT Centre for Robotics, School of Electrical
Engineering and Robotics, Brisbane, QLD 4000, Australia, and was partially supported by the QUT Centre for Robotics.
        {\tt\small \mmEM{}}}%
}

\begin{document}

\maketitle
\thispagestyle{empty}
\pagestyle{empty}

%%%%%%%%%%%%%%%%%%%%%%%%%%%%%%%%%%%%%%%%%%%%%%%%%%%%%%%%%%%%%%%%%%%%%%%%%%%%%%%%
\begin{abstract}
%VPR is a fundamental task for autonomous navigation as it enables a robot to localize itself in the workspace when a known location is detected.
%Although accuracy is an essential requirement for a VPR technique, computational and energy efficiency are not less important for real-world applications. CNN-based techniques archive state-of-the-art VPR performance, but they are computationally intensive and energy greedy. Binary Neural Networks (BNNs) have been recently proposed to address the VPR problem with significantly lower resource usage than CNNs.  Although a typical BNN is an order of magnitude faster than a CNN of similar size, its computational time can be further improved. In a typical BNN, the first convolution is not completely binarized for the sake of accuracy. Consequently, the first layer is the slowest network stage, spending a large share of the entire model's computational time. 
%This paper presents a BNN for VPR that combines \textit{depthwise separable factorization} and \textit{binarization} to replace the the first convolutional layer to improve computational and energy efficiency. 
%Our best model achieves \art{} VPR performance while spending considerably shorter time and energy than a BNN using a regular convolution as a first layer.
VPR is a fundamental task for autonomous navigation as it enables a robot to localize itself in the workspace when a known location is detected. Although accuracy is an essential requirement for a VPR technique, computational and energy efficiency are not less important for real-world applications. CNN-based techniques archive state-of-the-art VPR performance but are computationally intensive and energy demanding. Binary neural networks (BNN) have been recently proposed to address VPR efficiently. Although a typical BNN is an order of magnitude more efficient than a CNN, its processing time and energy usage can be further improved. In a typical BNN, the first convolution is not completely binarized for the sake of accuracy. Consequently, the first layer is the slowest network stage, requiring a large share of the entire computational effort. 
This paper presents a class of BNNs for VPR that combines \textit{depthwise separable factorization} and \textit{binarization} to replace the first convolutional layer to improve computational and energy efficiency. 
Our best model achieves \art{} VPR performance while spending considerably less time and energy to process an image than a BNN using a non-binary convolution as a first stage.

\end{abstract}

%\begin{IEEEkeywords}
%Performance Evaluation and Benchmarking; Visual-Based Navigation; Localization
%\end{IEEEkeywords}

%%%%%%%%%%%%%%%%%%%%%%%%%%%%%%%%%%%%%%%%%%%%%%%%%%%%%%%%%%%%%%%%%%%%%%%%%%%%%%%%
\section{Introduction}
\label{sec:intro}

Mobile robots need to track their position within the workspace to operate autonomously. As part of the navigation system, place recognition is fundamental in the localization process. It enables a robot to localize itself in the environment when a previously visited place on the map is detected. The rapid improvements in vision sensing capabilities made cameras the primary source of information for many robotics platforms motivating the interest in addressing place recognition with visual information \cite{maffra2019real,lowry2015visual}. Visual Place Recognition (VPR) primarily consists of matching the current camera view with an internal representation of the environment (a map) to determine the current robot's location. The changes occurring in an environment such as illumination, weather variations, and the different angles  from which the camera captures a place render VPR an arduous endeavor. Convolutional Neural Networks (CNNs) are successfully employed in VPR applications achieving \art{} performance under intense appearance changes \cite{zaffar2020vpr}.
However, CNNs have high runtime requirements unaffordable for many small robots \cite{maffra2018tolerant,ferrarini2019visual}. 
\begin{figure}[pt]
	\centering
	\includegraphics[width=0.97\columnwidth]{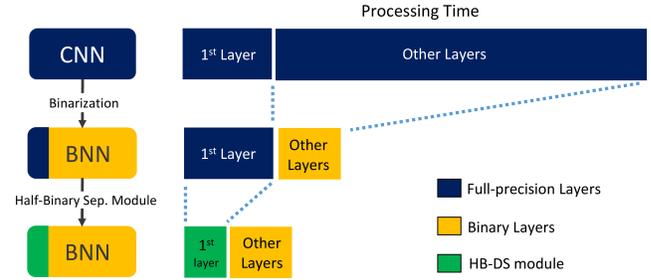}
	%\vspace{-3mm}
	\caption{In a typical BNN the first layer is not binarized for better accuracy. Hence, it is responsible for a significant part of the entire processing time. \HB{} module is a computationally efficient drop-in replacement to solve such a bottleneck problem in BNNs.}
\label{fig:fig1}
\end{figure}
Binary neural networks (BNN) \cite{courbariaux2016binarized} are a more efficient yet effective alternative to CNNs for enabling VPR in resource-constraint contexts %\cite{simons2019review,floppynet-pre}.
\cite{floppynet-pre}.
BNNs use a single bit to encode weights and activations, allowing compact model sizes and bitwise operations to achieve high computational efficiency \cite{hubara2017quantized}.  
Although a BNN can be one order of magnitude faster than a CNN, there is a substantial margin for improvement. For better classification and VPR accuracy, the first layer of BNNs takes high precision inputs \cite{hubara2017quantized,floppynet-pre}. Hence, the first convolution is incompatible with bitwise operations resulting in the most inefficient stage of a BNN, %\cite{bannink2020larq},
as exemplified in Fig. \ref{fig:fig1}.
%This is to better link my proposal to VPR. Intead of generally speaking of BNN, I specified why the first layer is a bottleneck.
%As the best image features for VPR are computed by the convolutional stege of a network for many enviromental changes \cite{hou2015convolutional,floppynet-pre}, the first layer is responsible for most of the latency of a BNN, as exemplified in Fig. \ref{fig:fig1}.
%The most effective image features for VPR are those computed by the convolutional stage of a network \cite{hou2015convolutional,hou2015convolutional,hou2015convolutional}. The same is for BNN as well \cite{floppynet-pre}. Consequently, the first layer might be responsible for most of the latency of a binary model, as exemplified in Fig. \ref{fig:fig1}.
Such a bottleneck problem is particularly relevant for VPR as many techniques use a relatively small number of convolutions \cite{floppynet-pre,Merrill2018RSS,bai2018sequence,chen2017only,tolias2016rmac}. Therefore, the first layer computes a significant part of the total operations to process an image that cannot be binarized without impacting the VPR performance.%, resulting responsible for most of the latency of a binary model.
%As the first convolution might be  Consequently, the first layer might be responsible for most of the latency of a binary model employed for VPR, as exemplified in Fig. \ref{fig:fig1}.
%
%
\begin{figure*}[pt]
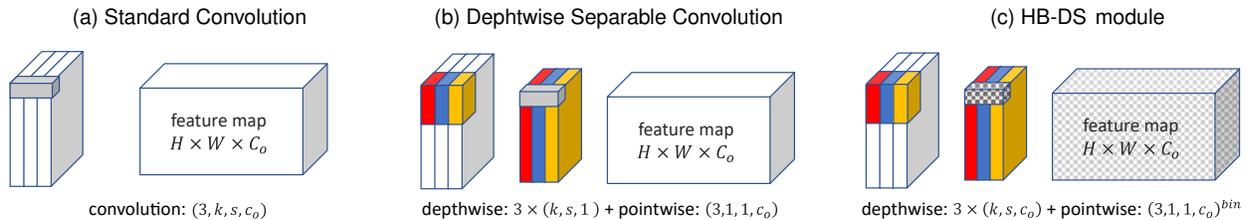

	\vspace*{1ex}
	\centering
	%\small\figtiTFlitefont{VPR matching performance comparison}\par\vspace*{3ex}
		\begin{annotatedFigure}{\includegraphics[width=0.31\linewidth]{\fullpath{half-binary3_a.pdf}}}
			\annotatedFigureText{0.27,0.99}{black}{0.0}{\footnotesize{(a)\;Standard\;Convolution}}
		\end{annotatedFigure}
		%First figure leads the positioning
		%\vspace{5ex}
		\hspace{1ex}
		\begin{annotatedFigure}{\includegraphics[width=0.31\linewidth]{\fullpath{half-binary3_b.pdf}}}
			\annotatedFigureText{0.07,0.99}{black}{0.0}{\footnotesize{(b)\;Dephtwise\;Separable\;Convolution}}
		\end{annotatedFigure}
		\hspace{1ex}
		\begin{annotatedFigure}{\includegraphics[width=0.31\linewidth]{\fullpath{half-binary3_c.pdf}}}
			\annotatedFigureText{0.34,0.99}{black}{0.15}{\footnotesize{(c)\;HB-DS module}}
		\end{annotatedFigure}	
\caption{A standard convolution works on all the inputs channels simultaneously to compute a feature map (a); depthwise separable factorization splits a convolution into depthwise and pointwise layers (b); half-binary separable convolution uses a binary pointwise convolution (c).}
\label{fig:separable}
\end{figure*}

This paper addresses the efficiency bottleneck of the first convolutional layer in BNNs by proposing the \textit{Half-Binary Depthwise Separable} (\HB{}) module. \HB{} combines depthwise separable factorization \cite{sifre2014rigid,howard2017mobilenets}
with binarization to enhance the computational efficiency of a BNN without affecting the VPR performance. 
The \HB{} module is then used to design a BNN that can be tuned to train models with several levels of efficiency to meet different application requirements.
%Our best model achieves \art{} performance requiring about \cTimeRatio{} of the time and energy spent by FloppyNet \cite{floppynet-pre} (a BNN for VPR), and $23\%$ spent by CALC, an efficient CNN for loop closure detection.
%
Our best model achieves \art{} performance requiring considerably lower resources than several VPR techniques regarded as highly efficient. For example, our network requires \cTimeRatio{} of the time and energy spent by FloppyNet \cite{floppynet-pre}, a BNN for VPR, and it is four times faster than CALC \cite{Merrill2018RSS}, a small-factor CNN proposed to address loop closure detection efficiently.

The rest of of this paper is organized as follows. Section \ref{sec:work} presents the related work. Section \ref{sec:method} describes the \HB{} module and the proposed network. The experimental setup and evaluation criteria are described in Section \ref{sec:setup}. Section \ref{sec:results} presents and discusses the experimental results. Conclusions are drawn in Section \ref{sec:conclusions}.

\section{Related Work}
\label{sec:work}

\subsection{Visual Place Recognition}
\label{sec:work:vpr}

Solving the VPR problem is the key to enable a robot to operate autonomously in a workspace. Despite the attention received and the significant advances in recent years, VPR remains challenging due to the viewpoint and appearance changes a robot encounters in real-world applications.
Among the most effective VPR techniques are those using CNNs, which achieve the highest performance in dynamic environments \cite{zaffar2020vpr,9459537,zaffar2019levelling}.    
The features computed by a CNN can be used as an image descriptor to match place images. A pre-trained AlexNet model \cite{krizhevsky2012imagenet} on Imagenet \cite{ILSVRC} is used for loop-closure detection \cite{7279659,bai2018sequence} and for enhancing the viewpoint tolerance of SeqSLAM \cite{milford2012seqslam,dongdong2018cnn}. AMOSNet and HybridNet \cite{chen2017deep} are CNN trained on Specific PlacEs Dataset (SPED) \cite{chen2017deep} with the aim to compute specific descriptors for place images. Zhou \etal{} followed the same idea proposing Place365 \cite{zhou2017places}, a large place dataset used to train several CNNs, including VGG-16 \cite{Simonyan14c}, to solve the VPR problem in changing environments. CNN features can be post-processed to compute a robust and compact image descriptor. R-MAC \cite{tolias2016rmac} applies a max pooling schema to the last convolutional layer of a pre-trained CNN and aggregates the resulting features into a vectorized image descriptor. Cross-Region-Bow \cite{chen2017only} and Regional-VLAD \cite{khaliq2018holistic} identify regions-of-interest (ROIs) in a pre-trained CNN's feature map and aggregate the underlying features using Bag-of-Visual-Words (BoW) \cite{philbin2007object} and VLAD \cite{jegou2010aggregating}, respectively. Unlike the others two-stage techniques, NetVLAD \cite{arandjelovic2016netvlad} trains the CNN and subsequent modules end-to-end to obtain a VLAD-like descriptor highly tolerant to viewpoint variations.

Although effective for VPR, CNNs are computationally intensive. CALC \cite{Merrill2018RSS} is an attempt to reduce the runtime requirement for VPR. It consists of a lightweight CNN trained using an autoencoder to recreate a HOG \cite{dalal2005histograms} descriptor from geometrically distorted place images. CoHOG \cite{zaffar2020cohog} is a trainingless methods proposed as a computationally efficient alternative to CNNs. It detects regions of interest using image-entropy \cite{memorable_maps} that are subsequently assigned with a HOG  descriptor to form an image representation.
%Recently solving the VPR problem in a efficient manner has received some attention.
Neurological-inspired techniques are considered as well to address VPR efficiently. FlyNet+CANN \cite{flynet} uses a compact pattern recognition stage followed by a time filter to match image sequences. DrosoNets \cite{drosonet} consists of an ensemble of compact bio-inspired place classifiers connected to a voting systems to determine the correct match with the camera's view.
Binary neural networks (BNNs) \cite{courbariaux2016binarized} use 1-bit parameters and bitwise arithmetic to speed-up convolution and reducing dramatically the memory usage. FloppyNet \cite{floppynet-pre} is a recently proposed BNNs for VPR applications. Its three-layer structure, along with bitwise arithmetic, makes FloppyNet a compact and computationally efficient image feature extractor. In this paper we propose a BNN with the same VPR accuracy as FloppyNet but substantially higher computational and energy efficiency. 

\subsection{Binary Neural Networks}
\label{sec:work:vpr:BNN}

CNNs successfully address the VPR problem but requires a heavy computational effort to build an image representation \cite{ferrarini2019visual,maffra2018tolerant}. Improving CNNs efficiency is a challenging task that received significant attention in the last decade. 
The earliest approaches reduce the computational complexity by pruning redundant connections and weights in trained models \cite{lecun1990optimal,hassibi1992second,han2015learning}. Post-training quantization is another method to reduce the computational requirements of a CNN. While post-quantization works reasonably well with eight or more bits in practical cases \cite{tfliteweb}, post-binarization affects the performance of a model dramatically \cite{courbariaux2014training}. %preso dalla 35 di XNOR
Binary-aware training is necessary to enable BNNs with good classification accuracy \cite{simons2019review}. Training a binary model was attempted decades ago \cite{saad1990training}.  However, only after introducing the Straight-Trough-Estimator (STE) method \cite{bengio2013estimating}, binary-aware training becomes easy to implement with gradient-based techniques. STE has become a standard in BNNs training. It is supported by an increasing number of machine-learning frameworks, including Larq \cite{geiger2020larq} and Brevitas \cite{brevitas}.

Since the first BNN trained with STE \cite{courbariaux2016binarized}, several additions to the field were made. 
%The accuracy was lower than AlexNet 64\% to 70\%, but the inference latency was about 20 times lower. Since Courbariaux's works there have been many addition to the field that improved BNNs contributing to their increasing popularity on restricted platforms. 
XNOR-Net \cite{rastegari2016xnor} places max-pooling layers before the quantization function to prevent pooling from binarized features that would result in a non-informative map overpopulated by \textit{ones}. Tang \etal{} \cite{tang2017train} learn a positive gain to apply to negative values to prevent frequent binary weight oscillations to ensure a shorter training and better performance. In \cite{esser2019learned} the binarization threshold is a learnable parameter for better classification accuracy.
%ABC-Net \cite{lin2017towards} approximate weights with a linear combinations of bases and uses multiple binary activations to alleviate the information loss.
The application of BNN to VPR is investigated in \cite{floppynet-pre}. Training a model using a fully connected stage including only full-precision neurons improves a model's performance when convolutional features are used for VPR.

An open problem in BNNs concerns the first convolutional layer, which is not binarized in most \art{} BNNs  to avoid performance loss \cite{simons2019review,alizadeh2018a}. Consequently, the first convolution is incompatible with bit-wise operations resulting in the slowest stage of a binary network (Section \ref{ssec:tuning}).  
%Most of BNNs are not completely binary. Some ancillary but necessary layers, such as Batch Normalizaion, are not binarizable. Others are kept full-precision to do not affect the performance. It is the case of the fist convolution. 
%The FBNA methodology \cite{8532584} consists of a module decomposing an input in a set of binary channels using multiple bases that are convolved separately and then recombined in a feature map. While effective, FBNA requires specifically FPGA designed modules for the deployment.  
%QuickNet \cite{bannink2020larq} addresses the first block latency using standard layers: a full-precision convolution with a few filters, 16, followed by a depthwise seprable convolutional block \cite{sifre2014rigid,howard2017mobilenets} to increase the feature map depth.
This paper proposes \HB{}, a module to replace the first convolutional layer to reduce the processing time of BNNs without affecting the VPR accuracy.

%\section{Half-Binary Depthwise Separable Module}
%\label{sec:method}

%%%
%
\begin{figure}[pt]
	\vspace{1ex}
	\centering
	\includegraphics[width=4.0cm, angle=270]{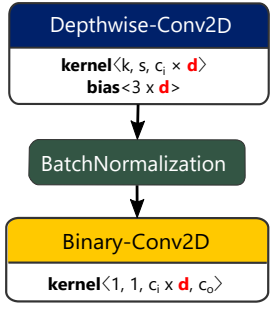}
	\vspace{1mm}
	\caption{HB-DS module implementation. $d$ denotes the depth multiplier, $k$ the kernel size, $s$ the stride, $c_i$ and $c_0$ the input and out channels, respectively.}
\label{fig:hb-block}
\end{figure}
%
%%%

\section{Solving the First Layer Bottleneck Problem}
%\section{A Solution to the First Layer Bottleneck Problem}
\label{sec:method}

The first convolutional layer is crucial for a BNN's performance
%Empirical evidence demonstrates that the encoding precision of the inputs plays an important role. 
A common practice in BNN design is using high-precision inputs because binarization negatively affects performance \cite{hubara2017quantized}. Consequently, the first convolutional layer is incompatible with bit-wise operations resulting in the slowest stage of  a BNN. For example, this is the case of FloppyNet \cite{floppynet-pre}, a recently proposed BNN designed for VPR applications consisting of three convolutions and three max-pool layers. The first FloppyNet's convolution takes approximately  $84\%$ of the total processing time, as detailed in Section \ref{ssec:tuning}.

This section presents the \textit{Half-Binary Depthwise Separable} module (\HB{}), to solve the first layer bottleneck problem. Then we use it to design a highly efficient BNN. The \HB{} module and the proposed BNN are detailed below.

\subsection{Depthwise Separable Convolutions}
Our approach uses depthwise separable factorization to split a convolution into two separate layers: a depthwise convolution and a pointwise convolution \cite{sifre2014rigid,howard2017mobilenets}. 
Depthwise convolution convolves the input channels individually. 
The input is convolved without changing the depth. Hence, the resulting feature map has the same channels as the input. The pointwise layer consists of a $1 \times 1$ convolution that builds a new map from the depthwise layer's features. Fig. \ref{fig:separable} shows the idea underlying depthwise separable decomposition.

%Our approach is based on depth-wise separable factorization that splits a convolution into two separate layers: depth-wise and point-wise convolutions.
%resulting about $k^2$ faster than a standard convolution with $k \times k$ kernel (Eq. \ref{eq:speed_factor}). 
The term complexity is used here as a synonym for the number of multiply-accumulate operations (MACs) computed by a convolution. Hence, lower computational complexity means fewer MACs.
Let us assume a convolutional layer takes an input tensor  $T_{in} = h_i \times w_i \times c_i$  and uses a kernel, $k \times k$, to output a feature map $T_{out} = h_o \times w_o \times c_o$. 
The computational cost is:
\begin{equation}
	C_{conv} = (k^2 \cdot c_i) \cdot h_o \cdot w_o \cdot c_o\text{\;,}
\end{equation}
%The parenthesis is to emphasize the computational cost for a single element in $T_{out}$.
%The term between parenthesis is the cost for a single location of $T_{out}$.
where $(k^2 \cdot c_i)$ is the cost for a single element in $T_{out}$.

%The first stage of a depthwise separable convolution, the depthwise convolution, convolves the input channels individually, creating a feature map having the same depth, $c_i$, as the input tensor. Figure  \ref{fig:separable}.b show the example of a color image as an input tensor.
A depthwise convolution convolves the input channels individually, creating a feature map having the same depth, $c_i$, as the input tensor. Fig. \ref{fig:separable}.b shows an example of a depthwise convolution processing a three-channel tensor (e.g. a color image).
%Depth-wise convolution can apply multiple kernels to an input channel creating a feature map with a depth of $d \cdot c_i$. $d$ denotes the \textit{depth multiplier} and it is part of the hyper-parameters of a network. 
The computational cost of a depthwise convolution is as follows:
\begin{equation}
	%C_{depth} = d \cdot ( k^2 \cdot c_{i}) \cdot h_{o} \cdot w_{o},  \;\;\;\;  d \in \mathbb{Z}_{>0}
	C_{depth} = k^2 \cdot c_{i} \cdot h_{o} \cdot w_{o}\text{\;.}
\end{equation}
The subsequent pointwise stage is a standard convolution with $k = 1$:
\begin{equation}	
	%C_{point} =  (d \cdot c_i) \cdot h_o \cdot w_o \cdot c_o,  \;\;\;\;  d \in \mathbb{Z}_{>0}
	C_{point} =  c_i \cdot h_o \cdot w_o \cdot c_o \text{\;,}
\end{equation}
The total computational cost of a depthwise separable convolution is:
\begin{equation}	
	%C_{sep} = c_i \cdot h_o \cdot w_o \cdot (k^2 + c_o) \cdot d
	C_{sep} = C_{depth} + C_{point} = c_i \cdot h_o \cdot w_o \cdot (k^2 + c_o)\text{\;.}
	\label{eq:csep}
\end{equation}
Compared to a standard convolution, the depthwise separable factorization reduces the complexity by:
\begin{equation}	
	%\frac{C_{sep}}{C_{conv}} = \left( \frac{1}{c_o} + \frac{1}{k^2}\right)\cdot d
	%\frac{C_{conv}}{C_{sep}} = d\cdot \frac{k^2 c_o}{c_o + k^2}
	\frac{C_{conv}}{C_{sep}} = \frac{k^2 c_o}{c_o + k^2}\text{\;.}
	\label{eq:sep_base_1}
\end{equation}
The larger the kernel, more effective is the depthwise separable factorization. 
%Nevertheless, the complexity of a convolution with a small kernel of $3 \times 3$ is reduced typically by a factor between $7$ and $9$.
%
%
%
%%%
\begin{figure}[pt]
	\vspace{1ex}
	\centering
	%\small\figtiTFlitefont{\footnotesize{Depthwise\;Separable\;Convolution\;($d=2$)}}
	\begin{annotatedFigure}{
		\includegraphics[width=7cm]{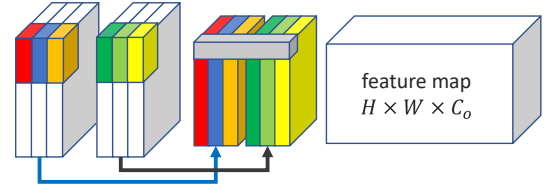}}
		%\annotatedFigureFilledText{0.65,0.38}{black}{0.2}{\scriptsize{feature\,map\\$H\times W \times C_{out}$}}{none}
	\end{annotatedFigure}
	\caption{Depthwise separable factorization with a depth multiplier of 2.}
	\label{fig:separable2}
\end{figure}
\subsection{Half-Binary Depthwise Separable Convolutions}
\label{ssec:bin_sep_conv}
We proposed a half-binary variant of depthwise separable factorization where only the second layer resulting from the decomposition is binarized (Fig. \ref{fig:separable}.c). Hence, the depthwise layer takes full precision inputs while the subsequent pointwise convolution is binary for higher computational efficiency.
The share of binary MACs is:
\begin{equation}	
	\frac{C_{point}}{C_{sep}} = \frac{c_o}{k^2 + c_o}\text{\;.}
	\label{eq:bin_share}
\end{equation}

Conversely, the full precision MAC are those in the depthwise convolution:
\begin{equation}	
	\frac{C_{depth}}{C_{sep}} = \frac{k^2}{k^2 + c_o}\text{\;.}
	\label{eq:fp_share}
\end{equation}
If $c_o > k^2$ the effect of binarization is dominant on factorization. Conversely, the complexity reduction is primarily due to factorization.

The implementation of the \HB{} module is illustrated in Fig. \ref{fig:hb-block}. A batch normalization layer \cite{batchnorm} is placed before the binary convolution to improve a model's accuracy and training speed
%\cite{alizadeh2018a,Sari2019HowDB,simons2019review}.  
\cite{alizadeh2018a}. 

%\subsection{Depth-Multiplier as a Tuning Parameter}
%\label{sec:depth-multi}

Depthwise convolution can apply multiple kernels to an input channel creating a thicker feature map. Let $d$ denotes the \textit{depth multiplier}. The resulting feature map has $d \cdot c_i$ channels, as exemplified in Fig. \ref{fig:separable2} for $d = 2$. The application of a depth multiplier increases the computational complexity of \HB{} by $d$ times. Therefore, Eq. \ref{eq:sep_base_1} is rewritten as follows:
%Equation \ref{eq:sep_base_1} is extended to include $d$ as follows:
\begin{equation}	
	%\frac{C^b_{sep}}{C_{conv}} = \left( \frac{1}{c_o} + \frac{1}{B \cdot k^2}\right)\cdot d
	\frac{C_{conv}}{C_{sep}} = \frac{k^2 c_o}{d(c_o + k^2)}\text{\;.}
\label{eq:with_d}
\end{equation}

On the other hand, the VPR performance of a BNN improves as $d$ increases. Section \ref{ssec:tuning} demonstrates the use of $d$ as a tuning parameter to adapt a model to different hardware capabilities while keeping \HB{} faster than an ordinary convolutional layer.

\begin{figure}[pt]
	\centering
	\vMargin
	\begin{annotatedFigure}{
		\includegraphics[width=0.6\columnwidth, angle=270]{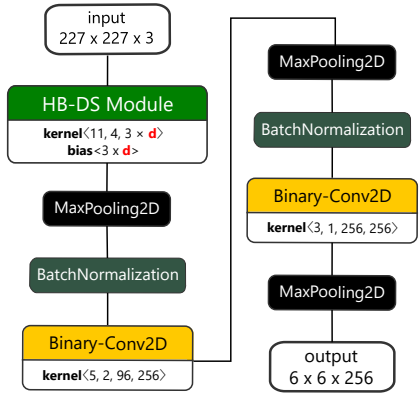}
		}
		%\annotatedFigureTextRotated{0.40,1.1}{black}{0.50}{\textrm{\normalsize{\HBFN{\textbf{-d}} Structure}}}{0}
	\end{annotatedFigure}
	\vspace{1mm}
	\caption{The proposed BNN uses \HB{} as a first stage. $d$ denotes the depth multiplier.}
\label{fig:bnn_arch}
\end{figure}

\subsection{Network Architecture}
\label{sec:architecture}

The proposed network shown in Fig. \ref{fig:bnn_arch} is inspired by FloppyNet. The \HB{} module uses a $11 \times 11$ kernel, stride of 4 and has 96 output channels. The rest of the network includes two pairs of binary convolution-max pooling blocks. The binary convolutions do not use bias but are preceded by batch normalization. The output features are from the last pooling layer. %and have $6 \times 6 \times 256 = 9216$ elements. %The number of parameters and MACs are shown in Tables \ref{tab:floppy} for various $d$ values.

\section{Esperimental Setup}
\label{sec:setup}

%Half-binary separable factorization is tested with two networks having different kernel sizes and output channels 
%in the first convolutional layer.
The proposed network is trained with several depth multipliers, $d$. The resulting models are assessed on VPR under various environmental changes. A model's efficiency is evaluated using processing time and energy usage as criteria.

\subsection{Training Data}
\label{sec:training_data}

All the binary models are trained from scratch using Place365 \cite{zhou2017places} within the Larq framework \cite{geiger2020larq}.
Places365 is a place-themed dataset consisting of 1,803,460 images divided into 365 classes, including between 3068 and 5000 samples. The validation set includes 100 images per category.

\subsection{Test Data}
\label{sec:datasets}

\begin{figure}[pt]
	\centering
	\vMargin
	\begin{annotatedFigure}{
			\includegraphics[width=0.97\columnwidth]{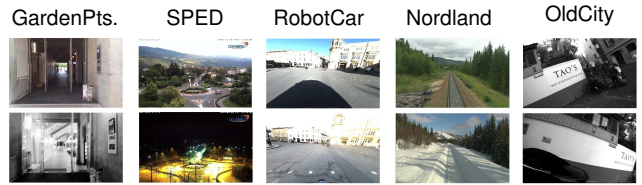}
		}
		\annotatedFigureText{0.005, 0.92}{black}{0.00}{\footnotesize{GardenPts.}}
		\annotatedFigureText{0.25, 0.92}{black}{0.00}{\footnotesize{SPED}}
		\annotatedFigureText{0.42, 0.92}{black}{0.00}{\footnotesize{RobotCar}}
		\annotatedFigureText{0.63, 0.92}{black}{0.00}{\footnotesize{Nordland}}
		\annotatedFigureText{0.85,0.92}{black}{0.00}{\footnotesize{OldCity}}
	\end{annotatedFigure}
	%\vspace{-5mm}
	\caption{A matching pair from every test dataset.}
\label{fig:dataset_example}
\end{figure}

VPR assessment is carried out under different image variations that a robot encounters over extended runs. The test data is divided into five datasets, each containing one or more image changes. They include: GardenPoints \cite{7353986}, 200 places randomly sampled from SPED \cite{chen2017deep}, the Cross-Season sequence from RobotCar \cite{larsson2019cross}, Nordlands \cite{nordlands} and Old City \cite{maffra2019real}. Table \ref{tab:datasets} summarizes the characteristics and the ground truth criteria for each dataset. All of them include a reference set representing the knowledge of the environment and a query set representing the current view of a robot's camera. Fig. \ref{fig:dataset_example} shows some examples of matching pairs.

%We included a sixth dataset, Combined, that results from the union of the \nDataset{} datasets mentioned above to simulate a large complex environment.It provides a more realistic global performance measure than the average of the \nDataset{} datasets tested separately.
We included a sixth dataset, Combined, to simulate a large complex environment. The reference set is the union of the other \nDataset{} datasets; the query set includes 200 randomly sampled images for a total of 1000 queries. The Combined dataset is intended to provide more realistic global performance measures than averaging the results from \nDataset{} datasets tested individually.

% Table generated by Excel2LaTeX from sheet 'dataset'
\begin{table}[pt]
  \centering
  \caption{Test datasets and ground truth tolerance.}
    \resizebox{0.97\columnwidth}{!}{
    \begin{tabular}{lcccc}
    \multicolumn{1}{l}{\multirow{2}[1]{*}{\textbf{Dataset}}} & \multirow{2}[1]{*}{\textbf{Condition}} & \multicolumn{1}{c}{\textbf{Reference}} & \multicolumn{1}{c}{\textbf{Query}} & \multicolumn{1}{c}{\textbf{Ground}} \\
          &       & \multicolumn{1}{c}{\textbf{Images}} & \multicolumn{1}{c}{\textbf{Images}} & \multicolumn{1}{c}{\textbf{Truth}} \bigstrut[b]\\
    \hline
    \hline
    \multirow{2}[2]{*}{GardenPoints} & Lateral Shift; & \multirow{2}[2]{*}{201} & \multirow{2}[2]{*}{201} & \multirow{2}[2]{*}{2 frames} \bigstrut[t]\\
          & Night-Day. &       &       &  \bigstrut[b]\\
    \hline
    \multicolumn{1}{l}{\multirow{2}[2]{*}{SPED}} & Weather; & \multirow{2}[2]{*}{1000} & \multirow{2}[2]{*}{200} & \multirow{2}[2]{*}{5 frames} \bigstrut[t]\\
          & \multicolumn{1}{c}{Night-Day.} &       &       &  \bigstrut[b]\\
    \hline
    \multirow{3}[2]{*}{\makecell[l]{RobotCar \\ Cross-Seasons}} & \multicolumn{1}{c}{Lateral Shift;} & \multicolumn{1}{c}{\multirow{3}[2]{*}{203}} & \multicolumn{1}{c}{\multirow{3}[2]{*}{180}} & \multicolumn{1}{c}{\multirow{3}[2]{*}{$\pm5$ frames}} \bigstrut[t]\\
    \multicolumn{1}{c}{} & \multicolumn{1}{c}{Ilumination;} &       &       &  \\
    \multicolumn{1}{c}{} & \multicolumn{1}{c}{Dynamic Elements.} &       &       &  \bigstrut[b]\\
    \hline

    \hline
    Nordland & \multicolumn{1}{c}{Seasons.} & 1622  & 1622  & 5 frames \bigstrut\\
    \hline
    Old City & \multicolumn{1}{c}{Strong 6-DOF.} & 5408  & 5643  & by authors \bigstrut[bt]\\
    \hline
    Combined & \multicolumn{1}{c}{All above.} & 8434  & 1000 & Mixed \bigstrut[bt]\\
    \hline
    \end{tabular}%
    }
  \label{tab:datasets}%
\end{table}%

\subsection{Evaluation Criteria}

\subsubsection{VPR Performance}
VPR is cast as a loop closure detection problem. A query image representing the current robot's camera view is compared to the reference images showing the previously visited locations. The image descriptor for our model is obtained from the vectorized output of a convolutional or pooling layer by L2-normalization:
\begin{equation}
    D = \frac{\hat{X_l}}{||\hat{X_l}||_2}\text{\;,}
    \label{eq:descr}
\end{equation}
where $\hat{X_l}$ is the output of the $l^{th}$ layer.
The similarity between the two images is determined using cosine:
\begin{equation}
    cos(\psi) = \frac{D_1\cdot D_2}{||D_1||\cdot||D_2||}\text{\;.}
\end{equation}
The reference image scoring the highest similarity with the query is regarded as the current location. 
VPR performance is measured on a dataset using several criteria including the percentage of true positive matches (TP\%) and two metrics computed from Precision-Recall curves:  Extended Precision (\perfS{}) \cite{EP} and AUC. \perfS{} extends the recall at $100\%$ precision ($\rp{}$) \cite{dongdong2018cnn} to the lower spectrum by incorporating the precision at the minimum recall ($\pr{}$):
\begin{equation}
\perfS{} = \frac{\pr{} + \rp{}}{2}\text{\;.} 
\label{eq:EP}
\end{equation}
%
%where $\rp{}$ denotes the recall at $100\%$ precision [ADD REF] and $\pr{}$ the precision at the minimum recall value. Differently from $\rp{}$ index, \perfS{} is able to determine the performance in the lower spectrum, namely in those cases where the precision does not reach $100\%$ at any recall.
%, namely where $\rp{}$ is not defined. 

\perfS{} is in $[0,1]$: higher the value, better the VPR performance. %It is worth mentioning that \perfS{} has the same meaning as $\rp{}$ in the interval $[0.5,1]$.

\subsubsection{Processing Time and Energy Usage}
\label{ssec:deply_metrics}

The processing time, $\tf{}$, and power usage, $\powerW{}$ are acquired from deployed models and techniques running on a test hardware platform.
$\tf{}$  is the time required  to elaborate an input image. The image loading and preprocessing (e.g. reshaping) are excluded so that $\tf{}$ reflects the actual computational complexity of a VPR technique.

The energy per image processed, $\energy{}$, indicates the energy spent to compute a single image representation.
It is determined from the power usage as follows: %$\energy{} = \power{}\tf{}$.

\begin{equation}
    \energy{} = \powerW{}\cdot\tf{}\text{,}
    \label{eq:energy}
\end{equation}
where $\powerW{}$ is the power absorbed during image processing.

\begin{figure}[pt]
	\vspace{1em}
	\centering
	%\textsc{\footnotesize\sffamily{FloppyNet's VPR matching performance}}\vspace*{1ex}
	%\includegraphics[width=0.97\columnwidth]{\fullpath{all_EP.pdf}}
	\includegraphics[width=0.97\columnwidth]{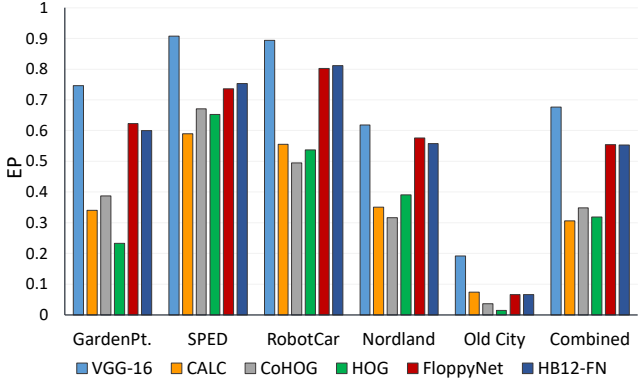}
	%\vspace{-5mm}
	\caption{VPR performance on different enviromental conditions.}
\label{fig:all_EP}
\end{figure}

\begin{figure}[pt]
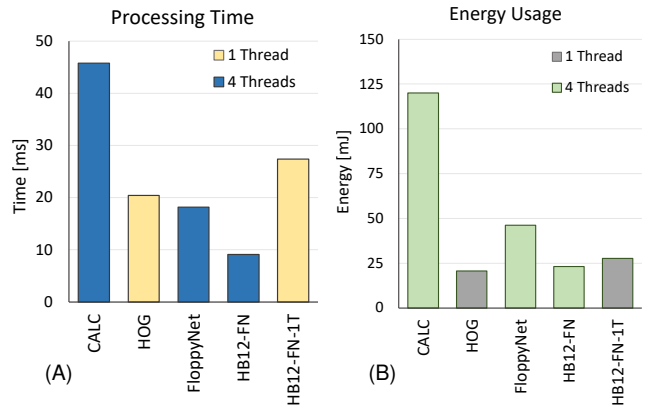

	\vspace{3em}
	\centering
	%\textsc{\footnotesize\sffamily{FloppyNet's VPR matching performance}}\vspace*{1ex}
	\begin{annotatedFigure}{
		\includegraphics[width=0.47\columnwidth]{\fullpath{all_time_12.pdf}}}
		\annotatedFigureText{0.1,0.05}{black}{0.00}{\footnotesize{(A)}}
	\end{annotatedFigure}
	\hspace{0mm}
	\begin{annotatedFigure}{
		\includegraphics[width=0.47\columnwidth]{\fullpath{all_energy_12.pdf}}}
		\annotatedFigureText{0.1,0.05}{black}{0.00}{\footnotesize{(B)}}
	\end{annotatedFigure}
	%\vspace{-5mm}
	\caption{Processing time (A) and energy usage (B) of the proposed BNNs compared to other VPR methods.}
\label{fig:all_time}
\end{figure}

%\section{Preliminary Results}
\section{Results Discussion}
\label{sec:results}

This section presents the results for the proposed BNN discussing both the VPR performance and efficiency.
%Depth multiplier, $d$, (Section \ref{ssec:bin_sep_conv}), controls the number of paramenters and MACs of \HB{} influencing the VPR performance and efficiency of a model.  
The experiments include several implementations of our BNN using different depth multipliers, $d$ (Section \ref{ssec:bin_sep_conv}). By convention, \HBFN{X} denotes a binary model using \HB{} with $d = X$. The computational time and energy usage are measured from deployed models. We used Larq-Compute-Engine (LCE) \cite{bannink2020larq} to run binary models on a Raspberry Pi4 (RPI4) \cite{rpi4-spec}. 

In the first part of this section, we compare \HBbest{} to other VPR techniques. We selected \HBbest{} as the best-balanced model between computation and performance. However, the experiments involved multiple models trained with different $d$ values. The second part of this section presents the relative results and discusses $d$ as a tuning parameter.

%We selected two models to compare with several VPR methods. \HBbest{} is best model in terms of VPR while \HBFN{12} has the same performance as FloppyNet \cite{floppynet-pre}, which is also included in the comparison, but it is more efficient. To the best of our knowledge, FloppyNet is the only BNN designed for VPR. Thus, it is considered as a baseline comparison for our BNNs. The other methods considered are VGG-16 \cite{Simonyan14c}, CALC \cite{Merrill2018RSS}, CoHOG \cite{zaffar2020cohog} and HOG \cite{dalal2005histograms}. 

% Table generated by Excel2LaTeX from sheet 'Tables'
\begin{table}[bp]
  \centering
  \caption{VPR metrics are given for the Combined Dataset. $\tf{}$ and $\energy{}$ are measured on a Raspberry PI4.}
\resizebox{0.97\columnwidth}{!}{
    \begin{tabular}{|l|l|c|c|ccc|}
    \hline
    \multicolumn{1}{|c|}{\multirow{2}[4]{*}{\textbf{VPR}}} & \multicolumn{1}{c|}{\multirow{2}[4]{*}{\textbf{Type}}} & \boldmath{}\textbf{$\boldsymbol{T_i}$}\unboldmath{} & \boldmath{}\textbf{$\boldsymbol{\energy{}}$ }\unboldmath{} & \multicolumn{3}{c|}{\textbf{VPR (combined)}} \bigstrut[t]\\
\cline{5-7}          &       & \textbf{ [ms]} & \textbf{[mJ]} & \textbf{EP} & \textbf{AUC} & \textbf{TP[\%]} \bigstrut[t]\\
    \hline
    \hline
    VGG-16 & CNN   & 995.7 & 2608.7 & 0.676 & 0.679 & 80.5 \bigstrut[t]\\
    CALC  & CNN   & 45.8  & 120   & 0.306 & 0.323 & 37.1 \\
    CoHOG & Trainless & 87.4  & 210.6 & 0.349 & 0.37  & 42.5 \\
    HOG   & Trainless & 20.4  & 20.6  & 0.318 & 0.335 & 38.6 \\
    FloppyNet & BNN   & 18.2  & 46.2  & 0.554 & 0.568 & 67.3 \\
%    \HBbest{} & BNN   & 10.9  & 27.7  & 0.569 & 0.587 & 68.1 \\
%    \HBbest{}-1T & BNN   & 33.3  & 33.6  & 0.569 & 0.587 & 68.1 \bigstrut[b]\\
    HB12-FN & BNN   & 9.1   & 23.1  & 0.553 & 0.566 & 67.2 \\
    HB12-FN-1T & BNN   & 27.4  & 27.7  & 0.553 & 0.566 & 67.2 \bigstrut[b]\\
    \hline
    \end{tabular}%
}
  \label{tab:all_tab}%
\end{table}%

% Table generated by Excel2LaTeX from sheet 'Tables'
\begin{table*}[tp]
  \centering
  \caption{Performance and efficiency for several implementation of the proposed BNN. $\tf{}$ and $\energy$ are measured on a Raspberry Pi 4.}
    \begin{tabular}{|l|c|c|cc|cc|cc|c|ccc|}
    \hline
    \multicolumn{1}{|c|}{\multirow{3}[6]{*}{\textbf{BNN}}} & \multicolumn{6}{c|}{\textbf{First Stage}}     & \multicolumn{2}{c|}{\multirow{2}[4]{*}{\boldmath{}\textbf{$\boldsymbol{T_i}$ [ms]}\unboldmath{}}} & \multirow{3}[6]{*}{\boldmath{}\textbf{$\boldsymbol{\energy{}}$ [mJ]}\unboldmath{}} & \multicolumn{3}{c|}{\textbf{VPR}} \bigstrut[t]\\
\cline{2-7}          & \multicolumn{1}{c|}{\textbf{Structure}} & \multirow{2}[4]{*}{\boldmath{}\textbf{$\boldsymbol{d}$ }\unboldmath{}} & \multicolumn{2}{c|}{\textbf{Params}} & \multicolumn{2}{c|}{\textbf{MAC [M]}} & \multicolumn{2}{c|}{} &       & \multicolumn{3}{c|}{(Combined Dataset)} \bigstrut[t]\\
\cline{4-9}\cline{11-13}          & \multicolumn{1}{c|}{\boldmath{}\textbf{($k$,$s$,$c_o$,$d$)}\unboldmath{}} &       & \multicolumn{1}{c|}{\textbf{32 bit}} & \textbf{1 bit} & \multicolumn{1}{c|}{\textbf{32bit}} & \textbf{1bit} & \multicolumn{1}{c|}{\textbf{First}} & \textbf{Total} &       & \multicolumn{1}{c|}{\boldmath{}\textbf{$\perfS{}$}\unboldmath{}} & \multicolumn{1}{c|}{\textbf{AUC}} & \textbf{TP (\%)} \bigstrut[t]\\
    \hline
    \hline
    HB1-FN & C(11,4,96) & 1     & 372   & 288   & 1.1   & 0.9   & 2.4   & 5.6   & 14.2  & 0.442 & 0.458 & 53.2 \bigstrut[t]\\
    HB4-FN & HD-BS(11,4,96,4) & 4     & 1488  & 1152  & 4.4   & 3.5   & 3.0   & 6.4   & 16.3  & 0.487 & 0.511 & 57.5 \\
    HB8-FN & HD-BS(11,4,96,8) & 8     & 2976  & 2304  & 8.8   & 7.0   & 4.8   & 8.2   & 20.8  & 0.554 & 0.566 & 67.8 \\
    HB12-FN & HD-BS(11,4,96,12) & 12    & 4464  & 3456  & 13.2  & 10.5  & 5.7   & 9.1   & 23.1  & 0.553 & 0.566 & 67.2 \\
    HB24-FN & HD-BS(11,4,96,24) & 24    & 8928  & 6912  & 26.4  & 20.9  & 7.7   & 10.9  & 27.7  & 0.569 & 0.587 & 68.1 \\
    HB48-FN & HD-BS(11,4,96,48) & 48    & 17856 & 13824 & 52.7  & 41.8  & 11.4  & 15.2  & 38.6  & 0.575 & 0.591 & 69.0 \\
    FloppyNet & HD-BS(11,4,96,N/A) & N/A   & 34848 & 0     & 105.4 & 0.0   & 15.1  & 18.2  & 46.2  & 0.554 & 0.568 & 67.3 \bigstrut[b]\\
    \hline
    \end{tabular}%
  \label{tab:floppy_table}%
\end{table*}%

\subsection{Comparative Analysis}
\label{sec:comparison}

The  VPR networks and methods included in the comparison are FloppyNet \cite{floppynet-pre}, VGG-16 \cite{Simonyan14c}, CALC \cite{Merrill2018RSS}, CoHOG \cite{zaffar2020cohog} and HOG \cite{dalal2005histograms}. To the best of our knowledge, FloppyNet is the only BNN designed for VPR. Thus, we consider it as a baseline comparison for \HBbest{}. 
VGG-16 is a large CNN used in several \art{} VPR applications \cite{arandjelovic2016netvlad,chen2017only,tolias2016rmac}. It includes 13 convolutions and computes about $15$M MACs to obtain an image representation from its last convolutional layer. The VGG-16 model used for the experiment is optimized on Place365. As opposed to VGG-16, CALC is a lightweight CNN designed to address VPR efficiently. We used the model trained on Place365 shared by the authors.  CoHOG is proposed as an efficient and training-less alternative to CNNs. For the experiments, we used the source code and the parameters shared by the authors for an image size of $256 \times 256$ pixels \cite{chog-repo}. Finally, HOG is used for $256 \times 256$ pixel images with a cell size of $16 \times 16$ and a block size of $32 \times 32$, as suggested in \cite{zaffar2020vpr}.

Fig. \ref{fig:all_EP} shows the EP score for all the considered methods. \HBbest{} and FloppyNet performs equally well on the Combined dataset while on GardenPoints and Nordalnd the latter achieves slight higher \perfS. Our network outperforms by a substantial margin the other lightweight techniques: HOG, CALC, and CoHOG. VGG-16 captures the highest EP score in every environmental condition and on the Combined dataset. These results are not surprising considering the VGG-16's large size and depth. However, VGG-16 requires $2.2s$ to compute an image descriptor, resulting in two orders of magnitude slower than any considered BNN. The complete set of $\tf{}$ is reported in Table \ref{tab:all_tab} while Fig. \ref{fig:all_time}.A compares a selection of the most efficient techniques: the BNNs, CALC, and HOG. 
%\HBbest{} and \HBFN{12} are the fastest ones taking $10.9\,ms$ and $9.1\,ms$ to process an image, respectively. In particular, \HBFN{12} scores comparable \perfS{} to FloppyNet while taking the $50\%$ of the time to complete an inference. 
\HBbest{} is the fastest one taking $\bestT{}$  to process an image, $\cTimeRatio{}$ of FloppyNet's inference time. Considering these two BNNs have similar VPR performance, our network represents a significant improvement over FloppyNet.
%In particular, \HBFN{12} scores comparable \perfS{} to FloppyNet while taking the $50\%$ of the time to complete an inference. 
The HOG implementation used for the experiments (OpenCV 4.5.0) can run only on a single thread. For a fair comparison, we reported the processing time of the proposed model for 1-thread (1T) execution using gray bars in Fig. \ref{fig:all_time}.A.
%HOG takes about $7\,ms$ less than \HBFN{12} but their VPR performance are on different levels. While HOG scores $0.365$, \HBFN{12} achieves $EP = 0.557$. Such a performance gap corresponds to a difference of $22.8\%$ less place correctly recognized on Combined dataset (Table \ref{tab:all_tab}).
HOG takes about $\HOGtimeGAP{}$ less than \HBbest{}-1T to compute a descriptor. However, their VPR performance is very different. While HOG scores $\perfS{} = \HOGEP{}$, \HBbest{} achieves $\bestEP{}$. Such a performance gap corresponds to $\HOGtpGAP{}$ less place correctly recognized in the Combined dataset (TP\% column in Table \ref{tab:all_tab}). We believe this gap is too wide to consider HOG as a good alternative to the proposed BNN.

\begin{figure}[pt]
	\centering
	%\textsc{\footnotesize\sffamily{FloppyNet's VPR matching performance}}\vspace*{1ex}
	\includegraphics[width=0.85\columnwidth]{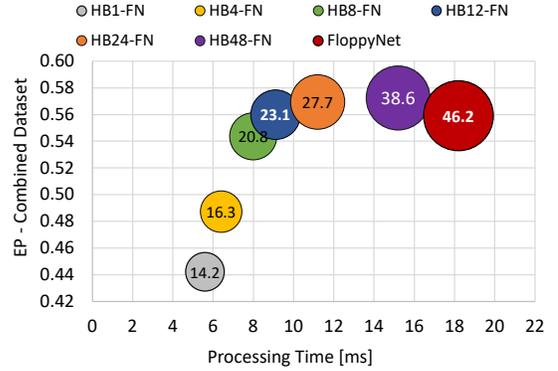}
	%\vspace{-5mm}
	\caption{EP score versus computational time for several depth multipliers. The circles' area represents the energy usage in $mJ$ per image processed.}
\label{fig:scatter_floppy}
\end{figure}

\subsection{Energy Usage}
\label{sec:energy_comparison}

The energy usage, $\energy$, is reported in Table \ref{tab:all_tab} and Fig. \ref{fig:all_time}.B. $\energy{}$ is determined using  Eq. \ref{eq:energy} from the average power usage measured directly from a RPI4 on 100 consecutive runs. 
RPI4 has an approximately continuous power usage during runtime. Thus, $\energy$ is mainly influenced by the interference time and the number of active CPU cores. To this end, the processing time reduction due to \HB{} contributes to energy saving, which is essential for battery-supplied robotic platforms. The positive effect of \HB{} is well depicted by the energy difference between FloppyNet and \HBbest{} as they differ only in the first stage. HOG is the most energy-efficient technique. However, as motivated above, HOG has too low VPR performance to replace our best model.
%The methods running on 1-thread (represented with gray bars) absorb less power as they activate only one core but take longer to complete. Indeed, running a model on a single thread does not provide any advantage in terms of energy usage as can be inferred by the similar $\energy{}$ values measured for
%%\HBFN{12} and \HBFN{12}-1T and for 
%\HBbest{} and \HBbest{}-1T. 
%Then, is possible comparing the energy usage of HOG, which runs on one thread, with \HBbest{}.

\begin{figure}[pt]
	\centering
	\vspace{4mm}
	%\textsc{\footnotesize\sffamily{FloppyNet's VPR matching performance}}\vspace*{1ex}
	\includegraphics[width=0.97\columnwidth]{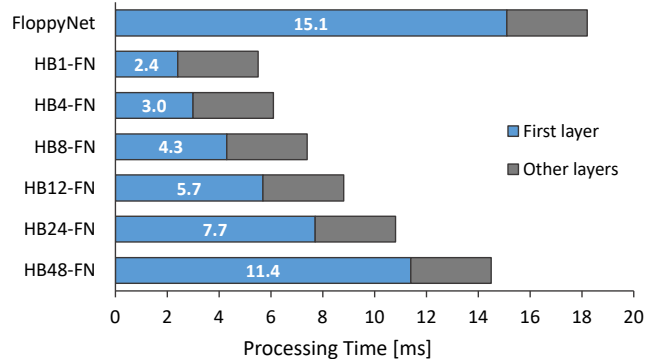}
	%\vspace{-5mm}
	\caption{Processing time for several depth multipliers. The blue bars represent the latency of the first layer while the gray bars indicate the rest of the network.}
\label{fig:floppy_time}
\end{figure}

\subsection{Depth Multiplier as a Tuning Parameter}
\label{ssec:tuning}

%https://www.universetoday.com/21293/despite-dust-storms-solar-power-is-best-for-mars-colonies/
%https://www.science.org/content/article/mammoth-dust-storm-mars-has-left-nasa-rover-dark

The \HB{} design enables the performance tuning of a BNN by acting only on the depth multiplier, $d$, without changing any other network parameter.
Fig. \ref{fig:scatter_floppy} plots the \perfS{} score on the Combined dataset versus the processing time for several depth multipliers. The circles' diameter represents the energy usage in $mJ$ per processed image. \HBbest{}, the model we selected for the comparison presented above, reaches the same VPR performance as FloppyNet, spending one-half of the processing time and energy. Nevertheless, the VPR performance can be further improved by increasing $d$. 
For example, \HBFN{24} and \HBFN{48} outperform \HBbest{} by a small margin at the cost of longer processing time and energy usage. If the target application has enough resources, \HBFN{24} and \HBFN{48} might be good options.
%resulted in the upper limit for the proposed BNN. Increasing the depth multiplier further extends the processing time and energy usage without any relevant benefits for VPR. The purple circle for $d = 48$ demonstrates the limit mentioned above. Another significant setup is $d = 12$ (blue circle). \HBFN{12} reaches the same \perfS{}, AUC and correct matches percentage as FloppyNet, spending one-half of the processing time and energy, as shown in Table \ref{tab:floppy_table}.
On the opposite side, lower $d$ values can find application in reducing a BNN's complexity to fit for extremely resource constraint hardware or saving energy to extend battery life.
For example, the \perfS{} loss from \HBbest{} to \HBFN{1} is $0.11$, which corresponds to $-14\%$ correctly matched places in the Combined dataset (TP\% in Table \ref{tab:floppy_table}). While \HBbest{} is possibly the best option in many scenarios, \HBFN{1} might be preferred when energy saving is a strict requirement as it spends less energy than \HBbest{} to process an image: $14.2\;mJ$ against $23.1\;mJ$. It is worth mentioning that \HBFN{1}, which holds the worst VPR performance among our models, outperforms HOG, CALC and CoHOG while achieving higher computational efficiency (Tables \ref{tab:all_tab} and \ref{tab:floppy_table}). Finally, Fig. \ref{fig:floppy_time} shows $\tf{}$ for several depth multipliers. The blue bars represents the time spent on the first layer, which depends on $d$. The gray bars are for the rest of the layers. FloppyNet use a regular non-binary convolution as a first stage. The time spent on the first convolution is $84\%$ of the entire processing time. The networks using \HB{} have a more fair distribution of the latency between the first and the other layers proving that our approach mitigates the bottleneck problem of BNNs.

\section{Conclusions}
\label{sec:conclusions}

BNNs are an efficient class of deep neural networks using binary arithmetic to speed up convolutions. The slowest stage in a BNN is the first convolutional layer, non-binary for higher accuracy. This paper proposed a BNN achieving \art{} VPR performance jointly with high computation and energy efficiency. Our network uses \HB{}, a module introduced in this paper, to address the latency bottleneck in the first stage of a BNN. \HB{} enables the performance tuning of a BNN by acting only on a single parameter to train models suitable for different deploy scenarios.
An extension of this work is investigating \HB{} for different network architectures (e.g. ResNet) and different tasks than image matching, such as object detection and image segmentation.

\bibliographystyle{abbrv}
\bibliography{bib}

\end{document}